# MedDoc-Bot: A Chat Tool for Comparative Analysis of Large Language Models in the Context of the Pediatric Hypertension Guideline

Mohamed Yaseen Jabarulla*, Steffen Oeltze-Jafra*, Philipp Beerbaum†, Theodor Uden†
*Peter L. Reichertz Institute for Medical Informatics of TU Braunschweig and Hannover Medical School,
†Department of Pediatric Cardiology and Pediatric Intensive Care Medicine, Hannover Medical School,
Hannover, Germany.
Email: Jabarulla.Mohamed@mh-hannover.de, Uden.Theodor@mh-hannover.de

*Abstract*— This research focuses on evaluating the non-commercial open-source large language models (LLMs) Meditron, MedAlpaca, Mistral, and Llama-2 for their efficacy in interpreting medical guidelines saved in PDF format. As a specific test scenario, we applied these models to the guidelines for hypertension in children and adolescents provided by the European Society of Cardiology (ESC). Leveraging Streamlit, a Python library, we developed a user-friendly medical document chatbot tool (MedDoc-Bot). This tool enables authorized users to upload PDF files and pose questions, generating interpretive responses from four locally stored LLMs. A pediatric expert provides a benchmark for evaluation by formulating questions and responses extracted from the ESC guidelines. The expert rates the model-generated responses based on their fidelity and relevance. Additionally, we evaluated the METEOR and chrF metric scores to assess the similarity of model responses to reference answers. Our study found that Llama-2 and Mistral performed well in metrics evaluation. However, Llama-2 was slower when dealing with text and tabular data. In our human evaluation, we observed that responses created by Mistral, Meditron, and Llama-2 exhibited reasonable fidelity and relevance. This study provides valuable insights into the strengths and limitations of LLMs for future developments in medical document interpretation. Open-Source Code: https://github.com/yaseen28/MedDoc-Bot

## I. INTRODUCTION

The clinical guidelines serve as crucial references for healthcare practitioners, guiding them in making informed decisions about patient care. Physicians not only in pediatric cardiology grapple with the task of assimilating and applying extensive guidelines [1], such as those related to hypertension in children and adolescents [2]. These medical and clinical practice guidelines vary between countries and healthcare associations [3]. The traditional manual practice involves laborious reading, which is time-consuming and susceptible to human error. Especially in emergency clinical scenarios, manually checking medical guidelines is impractical.

Moreover, with the proliferation of general-purpose language models [4], it may seem plausible to leverage them for medical guidelines interpretation. However, existing models often fall short of specificity, struggling to provide accurate answers to clinically relevant questions within guidelines. Limited access to the underlying language model makes it difficult to utilize closed-source models like PaLM and GPT-4 for domain-specific applications, such as analyzing sensitive patient records. In contrast, open-source models can be executed on local computers, allowing sensitive patient information to be entered [5], [6].

Over the past years, researchers have developed medical domain-specific LLMs such as Med-Palm [7], PMC-LLaMA [8], and Clinical Camel [9] to assist doctors in automating tasks, including summarizing medical records and extracting relevant information from PubMed open-access research papers [10]. Notably, recent task-specific models like Meditron, MedAlpaca, and ChatDoctor [5], [6] further improve LLMs' interpreting and rational reasoning capabilities by continuous pre-training on general or domain-specific open source models [11], [12] using carefully curated datasets. Thus, the model generates a better response to task-specific prompts by utilizing the combination of natural and domain-specific language [9].

However, these models face challenges in real-world clinical settings, especially when dealing with detailed and more specific medical guidelines. For instance, querying a guideline for precise information on medication dosage or treatment protocols for children requires a level of specificity that most broad-scoped models lack. The inadequacy of existing models in catering to these needs hinders language models' seamless integration into pediatric cardiologists' daily workflows. Despite the availability of numerous language models, we acknowledge the challenges faced by healthcare experts in seamlessly incorporating and testing these models in real-time clinical settings.

In light of these challenges, our research aims to bridge this gap by developing a PDF chat tool that harnesses the power of four new quantized LLMs: Meditron, MedAlpaca, Mistral, and Llama-2. The quantized variants from Hugging Face are used to ensure a balance between reduced model size and performance, making the system well-suited for local implementations with limited resources. On the front end, a user-friendly multi-PDF chatbot tool (MedDoc-Bot) was constructed with Streamlit V1.30 [13], a Python-based web application tool for machine learning and data science. This enables users to upload single or multiple PDF documents to the interface. A single prompt can trigger a search for relevant information across all uploaded PDFs and provide responses based on the selected LLMs' capabilities. The backend utilizes LangChain—a framework designed for language-driven applications for document processing [14]. In our clinical application, we evaluated the performance of each model by interpreting the ESC hypertension guidelines for children and adolescents [2]. We prompted each model with a pediatric expert-curated benchmark dataset. Our evaluation criteria included assessing the models' response fidelity, relevance, chrF (Character n-gram F-score), and METEOR (Metric for Evaluation of Translation with Explicit Ordering) scores [15], [16], providing insights into the correctness, linguistic quality, and overall performance of their responses when queried about pediatric hypertension guidelines.

## II. Materials and methods

### A. Open-source Large Language Models

For our study, we chose two general-purpose models, Llama-2 and Mistral 7B, as well as two medical domain-specific models, Meditron, and MedAlpaca to prompt queries in the context of ESC pediatric hypertension guidelines. The key features and the quantized version utilized in our research study are discussed below.

LLAMA-2 is part of state-of-the-art LLMs with 7 billion to 70 billion parameters [12]. LLAMA-2 is optimized for dialogue, surpassing open-source chat models in benchmarks. It utilizes an auto-regressive transformer architecture and undergoes supervised fine-tuning and reinforcement learning with human feedback. Our evaluation study focused on Llama-2 13B, a 13-billion-parameter variant. Mistral 7B [11] is an advanced generative text model that leverages grouped-query attention and sliding window attention for faster inference and effective handling of sequences of arbitrary length with reduced inference costs. Mistral 7B outperforms other models across various benchmarks, including reasoning, mathematics, and code generation. We utilized Mistral 7B Instruct V0.2, which is a fine-tuned version of Mistral 7B and Mistral-7B-v0.1 that uses a variety of publicly available conversation datasets.

Meditron-7B [5], specifically designed for the medical domain with 7 billion parameters, surpasses other models in medical reasoning tasks. It is pre-trained on curated medical data to assist clinical decision-making and improve access to LLMs in healthcare. The MedAlpaca-13B [6] focuses on improving medical question-answering and dialogue tasks. It is a fine-tuned LLaMa model trained on diverse datasets, including Anki flashcards, Wikidoc, and StackExchange.

The four models are pre-quantized in GGUF (GPT-Generated Unified Format) format introduced by the LLaMA C++ community and retrieved from the Hugging Face repository. This strategic pre-quantization facilitates efficient processing to accommodate diverse local computer setups, considering potential CPU or GPU limitations. This ensures that the models can be effectively utilized across various computing environments, striking a harmonious equilibrium between computational efficiency and high-performance analysis.

### B. Dataset and Clinical Use Case

In this section, we delve into the critical aspects of dataset preparation and a clinical use case to evaluate the effectiveness of language models using our PDF chat tool interface in the context of pediatric hypertension guidelines.

To evaluate the efficacy of four LLMs, a pediatric specialist with four years of experience in pediatric cardiology manually generated twelve questions and corresponding responses by meticulously reviewing the pediatric hypertension guidelines. This dataset, serving as a benchmark, is divided into three groups: (1) *Clinical*: Questions related to clinical scenarios or medical cases. (2) *Visual Element*: Questions derived from tables and figures in the dataset. (3) *General*: General inquiries, such as definitions or background information, extracted from the guideline´s content. Each group, containing four questions and responses, ensures a thorough evaluation of each model's interpretative capabilities using the uploaded PDF guideline document in the chat tool. Table 1 illustrates an exemplary question and its corresponding response derived from the clinical questions group. These questions are accessible from our Github repository.

TABLE 1: Example from Clinical Questions Group and Corresponding Response Generated by Pediatric Specialist.

| |
|---|
| **Question:** What is the proposed cut-point for identifying left ventricular hypertrophy (LVH) by echocardiography in children? |
| **Response:** The proposed cut-point for identifying LVH by echocardiography in this age range is ≥45 g/m^2. Alternatively, LVH may also be defined by the 95th percentile of height normalized for age and sex. |

The pediatric hypertension guideline [2] contains text, tables, and figures on twelve pages. We carefully transformed figures and tables into textual representations to enhance interpretation and extraction. This involves providing detailed captions, extracting numerical data, and describing visual features in text. The transformed PDF file is uploaded along with the original online version for evaluating the response generated from *visual element* queries. These documents collectively support assessing the language model's ability to respond to queries about visual elements within the guidelines (including figures, tables and their textual representation).

### C. Methods

This section describes the methodology employed for our Streamlit-powered MedDoc-Bot chatbot. The pre-quantized LLMs' Meditron, MedAlpaca, Mistral, and Llama-2 are stored locally on our machine. We initialized these state-of-the-art LLMs using CTransformers, a Python library based on the Transformer architecture, contributing to efficient computations in CPU or mixed CPU/GPU environments.

As illustrated in Fig. 1, the authenticated user-uploaded PDF document undergoes preprocessing utilizing LangChain. The LangChain captures relevant information from the hypertension guideline [2] document by transforming large texts into smaller chunks. In the subsequent phase, the processed chunks are transformed into numerical vectors that represent the semantic meaning of the text. This facilitates efficient identification of similar texts in the vector space, which are then stored in vector databases. In this study, the state-of-the-art sentence transformers [17] framework is used to embed text chunks into numerical representations for constructing a semantic index in a vector database (DB). FAISS [18] library is utilized to store embeddings in a vector DB, ensuring a quick semantic search for information extracted from PDF documents.

User query processing incorporates Sentence Transformers to embed queries into numerical representations. Semantic searches are executed in the vector store database, retrieving relevant information for a personalized user experience. LLMs, such as Meditron or MedAlpaca, generate responses based on contextual relevance, fidelity, and consistency, thereby enhancing information retrieval precision. We highlight that the Streamlit-based user interface provides users with the option of selecting from multiple language models for document processing and seamlessly uploading pediatric hypertension guidelines in PDF format. Streamlit's user-friendly design enhances accessibility, providing an interface aligned with streamlining multi-pdf interpretation.

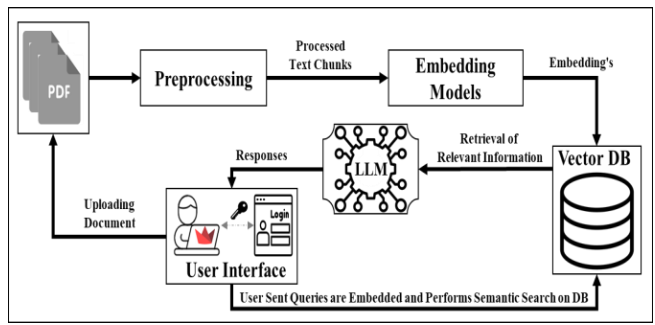

Fig. 1. Overview of the StreamLit-powered MedDoc-Bot chat tool process.

### III. EXPERIMENTAL RESULTS

We utilized Python to implement the MedDoc-Bot for evaluating four language models on our local machine. Our system is equipped with a 12th Gen Intel i9 core processor, 64 GB of RAM, and an NVIDIA GeForce RTX 3060 graphics card. In this experimental analysis, we categorized the benchmark dataset into three groups (see section II(B)). Evaluating the models' responses involved human verification and metrics like chrF and METEOR for queries related to both textual and visual elements from the uploaded PDF guideline. Throughout the evaluation process, questions were put into the MedDoc-Bot interface, prompting each model to generate responses. Posing the same questions to all tested models allowed us to compare consistency, fidelity, and relevance of answers between models. Additionally, we recorded the time taken by each model to generate responses to each query.

We conducted a comprehensive dual assessment, evaluating the human relevance of the generated responses. This involved assessing their meaningfulness and accuracy in addressing medical queries and their fidelity to the original guideline text. A pediatric expert provided ratings on a scale from 0 to 100% (ranging from entirely irrelevant to fully addressing the question/fitting the guideline text), and Fig. 2 presents the consolidated average accuracy scores for four language models. Notably, Mistral and Llama-2 excelled in both relevance and fidelity. Meditron performed moderately, while MedAlpaca exhibited lower scores in the benchmark dataset, indicating poor generalizability.

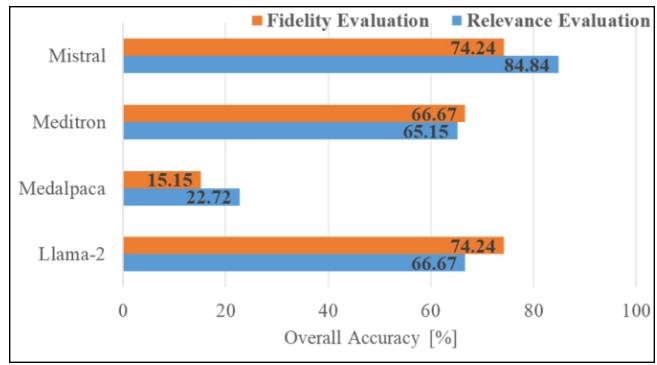

Fig.2. Human expert-evaluated average accuracy of responses from four language models. We comprehensively assessed model-generated responses based on relevance and fidelity to the original guideline text.

Our assessment criteria encompass not only responses accuracy but also linguistic similarity aspects, employing metrics such as chrF and METEOR Scores. chrF [15] is a metric that operates at the character level, measuring the similarity between LLMs' generated texts and reference texts. The METEOR score [16] is another metric for evaluating machine-generated text by analysing overall quality, considering precision, recall, and penalties.

Table 2 provides a comprehensive evaluation of LLMs across three benchmark datasets: clinical, visual, and general. Each dataset includes four expert-created questions and corresponding reference responses, showcasing LLM performance variation. Based on table 2, Llama-2 leads in the *clinical dataset (G1)*, followed by Mistral and Meditron, while MedAlpaca shows lower average scores. Similar patterns are observed in the *visual element (G2)* and the *general dataset (G3)*. Across all datasets, Mistral and Meditron maintain competitive performance. Conversely, MedAlpaca consistently lags behind, suggesting potential limitations in generating high-quality responses. This analysis underscores Llama's reliability across diverse contexts, positioning it as a strong choice for applications prioritizing response quality. Mistral also proves consistently effective, though MedAlpaca and Meditron may benefit from improvements. The findings highlight nuanced performance variations among LLMs and emphasize the critical balance between quality and efficiency in response generation across different benchmark datasets. The chrF and METEOR scores, ranging from 0 to 1, provide a baseline match. This could be due to the fact that open source LLMs were not yet fine-tuned for pediatric guidelines. The applied metrics mostly consider the similarity between reference and language model responses. However, the human expert evaluation considers that answers might be very accurate and relevant, even though formulated in completely different words than the reference responses.

Table 2: Average METEOR and chrF scores provide a consolidated performance overview. Scores are averaged across clinical (G1), visual elements (G2), and general (G3) response datasets group for four LLMs.

| Group | Llama-2 Score | | MedAlpaca Score | | Meditron Score | | Mistral Score | |
|---|---|---|---|---|---|---|---|---|
| | Meteor | chrF | Meteor | chrF | Meteor | chrF | Meteor | chrF |
| G1 | **0.50** | **0.53** | 0.21 | 0.33 | 0.41 | 0.47 | 0.46 | 0.48 |
| G2 | **0.32** | **0.42** | 0.10 | 0.18 | 0.20 | 0.33 | 0.21 | 0.32 |
| G3 | 0.23 | 0.34 | 0.18 | 0.29 | 0.32 | 0.42 | **0.34** | **0.43** |

Fig. 3 depicts the average time taken (in minutes) by each LLMs to create responses across three group datasets. We can observe that for *clinical* questions, Meditron and Mistral are faster, compared with longer times for MedAlpaca and Llama-2. Mistral stands out with the fastest response time for *visual elements* questions. For *general* questions, Llama-2 takes the longest, while MedAlpaca is notably quicker. Overall, MedAlpaca and Mistral generally show efficiency, while Llama-2's response times are longer, potentially reflecting trade-offs between response quality and speed. These time metrics offer crucial insights for applications emphasizing timely outputs and influence model selection based on specific performance needs.

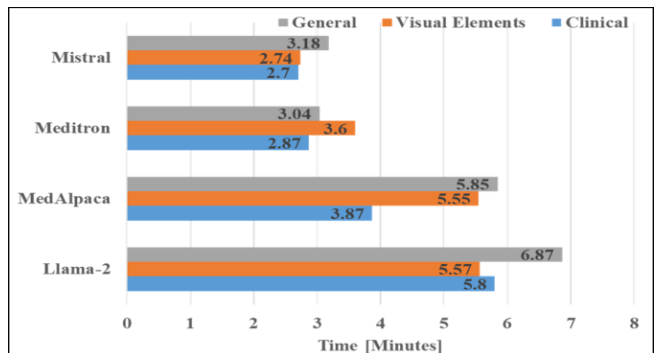

Fig. 3. Average Response Time (in minutes) of LLMs' across datasets.

## IV. CONCLUSION

In this preliminary research, we analyzed and evaluated four LLMs - Meditron, MedAlpaca, Mistral, and Llama-2 - for their ability to interpret pediatric guidelines. Streamlit was used to develop a MedDoc-Bot chat tool for locally interacting with these models. This tool allows authorized users to upload PDF files and ask questions. The evaluation criteria, encompassing fidelity, relevance, chrF, and METEOR scores, are crucial for assessing correctness and linguistic quality in LLM-generated responses related to pediatric hypertension guidelines, especially in the context of visual information processing. While our focus is on pediatric guidelines, we acknowledge the tool's versatility for other PDF documents. A human evaluation finds Mistral, Llama-2, and Meditron to be reliable and relevant based on their interpretation of pediatric guideline. Comparative analysis reveals Llama-2 excels with the best METEOR and chrF scores, especially in clinical responses, while Mistral performs satisfactorily. The MedAlpaca and Meditron systems consistently lag behind, suggesting possible limitations. Ongoing work involves fine-tuning the best-performing model (Llama-2 and Mistral) with a clinical dataset curated by multiple experts for secure patient record analysis on a local system.